\documentclass[11pt,letterpaper]{article}
\usepackage{naaclhlt2016}
\usepackage{times}
\usepackage{url}
\usepackage{latexsym}

\usepackage{yfonts}
\usepackage{graphicx} 
\usepackage{subfigure}

\usepackage[linesnumbered,ruled,vlined]{algorithm2e}

\SetCommentSty{mycommfont}

\usepackage{verbatim}
\usepackage{etoolbox}
\usepackage{framed}
\usepackage{xcolor}
\usepackage{color}
\usepackage{amsmath}
\newcommand{\argmax}{\operatornamewithlimits{argmax}}

\usepackage{multirow}
\usepackage{arydshln}

\naaclfinalcopy 
\def\naaclpaperid{123} 

\title{Unsupervised Ranking Model for Entity Coreference Resolution}

 \author{Xuezhe Ma \and Zhengzhong Liu \and Eduard Hovy \\
Language Technologies Institute \\
Carnegie Mellon University \\
Pittsburgh, PA 15213, USA \\
{\tt \{xuezhem, liu\}@cs.cmu.edu, ehovy@cmu.edu}}

\date{}

\begin{document}

\maketitle

\begin{abstract}
Coreference resolution is one of the first stages in deep language understanding and its importance has been well recognized in the natural language processing community. In this paper, we propose a generative, unsupervised ranking model for entity coreference resolution by introducing resolution mode variables. Our unsupervised system achieves 58.44\% F1 score of the CoNLL metric on the English data from the CoNLL-2012 shared task~\cite{pradhan-EtAl:2012:CoNLL-2012-ST}, outperforming the Stanford deterministic system~\cite{Lee:2013:CL} by 3.01\%.
\end{abstract}

\section{Introduction}
Entity coreference resolution has become a critical component for many Natural Language Processing (NLP) tasks. Systems requiring deep language understanding, such as information extraction~\cite{wellner2004integrated}, semantic event learning~\cite{chambers-jurafsky:2008:ACLMain,chambers-jurafsky:2009:ACLIJCNLP}, and named entity linking~\cite{DurrettKlein2014,ji2014overview} all benefit from entity coreference information.

Entity coreference resolution is the task of identifying mentions (i.e., noun phrases) in a text or dialogue that refer to the same real-world entities. In recent years, several supervised entity coreference resolution systems have been proposed, which, according to \newcite{ng:2010:ACL}, can be categorized into three classes --- mention-pair models~\cite{mccarthy1995using}, entity-mention models~\cite{yang2008entity,haghighi-klein:2010:NAACLHLT,lee-EtAl:2011:CoNLL-ST} and ranking models~\cite{yang2008twin,durrett-klein:2013:EMNLP,fernandes2014latent} --- among which ranking models recently obtained state-of-the-art performance. However, the manually annotated corpora that these systems rely on are highly expensive to create, in particular when we want to build data for resource-poor languages~\cite{ma-xia:2014:P14-1}. That makes unsupervised approaches, which only require unannotated text for training, a desirable solution to this problem.

Several unsupervised learning algorithms have been applied to coreference resolution. \newcite{haghighi-klein:2007:ACLMain} presented a mention-pair nonparametric fully-generative Bayesian model for unsupervised coreference resolution. Based on this model, \newcite{ng:2008:EMNLP} probabilistically induced coreference partitions via EM clustering. \newcite{poon-domingos:2008:EMNLP} proposed an entity-mention model that is able to perform joint inference across mentions by using Markov Logic. Unfortunately, these unsupervised systems' performance on accuracy significantly falls behind those of supervised systems, and are even worse than the deterministic rule-based systems. Furthermore, there is no previous work exploring the possibility of developing an unsupervised ranking model which achieved state-of-the-art performance under supervised settings for entity coreference resolution.

In this paper, we propose an unsupervised generative ranking model for entity coreference resolution. Our experimental results on the English data from the CoNLL-2012 shared task~\cite{pradhan-EtAl:2012:CoNLL-2012-ST} show that our unsupervised system outperforms the Stanford deterministic system~\cite{Lee:2013:CL} by 3.01\% absolute on the CoNLL official metric. The contributions of this work are (i) proposing the first unsupervised ranking model for entity coreference resolution. (ii) giving empirical evaluations of this model on benchmark data sets. (iii) considerably narrowing the gap to supervised coreference resolution accuracy.

\section{Unsupervised Ranking Model}
\label{sec:model}
\subsection{Notations and Definitions}
In the following, $D = \{m_0, m_1, \ldots, m_n\}$ represents a generic input document which is a sequence of coreference mentions, including the artificial root mention (denoted by $m_0$). The method to detect and extract these mentions is discussed later in Section~\ref{subsec:mention-detect}. Let $C = \{c_1, c_2, \ldots, c_n\}$ denote the coreference assignment of a given document, where each mention $m_i$ has an associated random variable $c_i$ taking values in the set $\{0, i, \ldots, i-1\}$; this variable specifies $m_i$'s selected antecedent ($c_i \in \{1, 2, \ldots, i-1\}$), or indicates that it begins a new coreference chain ($c_i = 0$).

\subsection{Generative Ranking Model}
The following is a straightforward way to build a generative model for coreference:
\begin{equation}\label{eq:model}
\begin{array}{rcl}
P(D, C) & = & P(D|C)P(C) \\
& = & \prod\limits_{j=1}^{n}P(m_j|m_{c_j})\prod\limits_{j=1}^{n}P(c_j|j)
\end{array}
\end{equation}
where we factorize the probabilities $P(D|C)$ and $P(C)$ into each position $j$ by adopting appropriate independence assumptions that given the coreference assignment $c_j$ and corresponding coreferent mention $m_{c_j}$, the mention $m_j$ is independent with other mentions in front of it. This independent assumption is similar to that in the IBM 1 model on machine translation~\cite{brown1993mathematics}, where it assumes that given the corresponding English word, the aligned foreign word is independent with other English and foreign words. We do not make any independent assumptions among different features~(see Section~\ref{subsec:feats} for details).

Inference in this model is efficient, because we can compute $c_j$ separately for each mention:
\begin{displaymath}
c^*_j = \argmax\limits_{c_j} P(m_j|m_{c_j}) P(c_j|j)
\end{displaymath}
The model is a so-called ranking model because it is able to identify the most probable candidate antecedent given a mention to be resolved.

\begin{table*}[t]
\centering
{\scriptsize
\begin{tabular}[t]{c|c|l}
\hline
Mode $\pi$ & Feature & Description \\
\hline
\emph{prec} & Mention Type & the type of a mention. We use three mention types: $Proper, Nominal, Pronoun$ \\
\hline
\multirow{4}{*}{\emph{str}} & Mention Type & the same as the mention type feature under \emph{prec} mode. \\
 \cline{2-3}
 & Exact Match & boolean feature corresponding to String Match sieve in Stanford system. \\
 \cline{2-3}
 & Relaxed Match & boolean feature corresponding to Relaxed String Match sieve in Stanford system.\\
 \cline{2-3}
 & Head Match & boolean feature corresponding to Strict Head Match A sieve in Stanford system. \\
\hline
\multirow{7}{*}{\emph{attr}} & Mention Type & the same as the mention type feature under \emph{prec} mode. \\
 \cline{2-3}
 & Number & the number of a mention similarly derived from \newcite{Lee:2013:CL}. \\
 \cline{2-3}
 & Gender & the gender of a mention from \newcite{bergsma-lin:2006:COLACL} and \newcite{ji2009gender}. \\
 \cline{2-3}
 & Person & the person attribute from \newcite{Lee:2013:CL}. We assign person attributes to all mentions, not only pronouns. \\
 \cline{2-3}
 & Animacy & the animacy attribute same as \newcite{Lee:2013:CL}. \\
 \cline{2-3}
 & Semantic Class & semantic classes derived from WordNet~\cite{soon2001machine}. \\
 \cline{2-3}
 & Distance & sentence distance between the two mentions. This feature is for parameter $q(k|j, \pi)$ \\
\hline
\end{tabular}
}
\caption{Feature set for representing a mention under different resolution modes. The \emph{Distance} feature is for parameter $q$, while all other features are for parameter $t$.}
\label{tab:feats}
\end{table*}

\subsection{Resolution Mode Variables}
According to previous work~\cite{haghighi-klein:2009:EMNLP,ratinov-roth:2012:EMNLP-CoNLL,Lee:2013:CL}, antecedents are resolved by different categories of information for different mentions. For example, the Stanford system ~\cite{Lee:2013:CL} uses string-matching sieves to link two mentions with similar text and precise-construct sieve to link two mentions which satisfy special syntactic or semantic relations such as apposition or acronym. Motivated by this, we introduce resolution mode variables $\Pi = \{\pi_1, \ldots, \pi_n\}$, where for each mention $j$ the variable $\pi_j \in \{str, prec, attr\}$ indicates in which mode the mention should be resolved. In our model, we define three resolution modes --- string-matching (\emph{str}), precise-construct (\emph{prec}), and attribute-matching (\emph{attr}) --- and $\Pi$ is deterministic when $D$ is given (i.e. $P(\Pi|D)$ is a point distribution). We determine $\pi_j$ for each mention $m_j$ in the following way:
\begin{itemize}
\item $\pi_j = str$, if there exists a mention $m_i, i < j$ such that the two mentions satisfy the \emph{String Match} sieve, the \emph{Relaxed String Match} sieve, or the \emph{Strict Head Match A} sieve in the Stanford multi-sieve system~\cite{Lee:2013:CL}.
\item $\pi_j = prec$, if there exists a mention $m_i, i < j$ such that the two mentions satisfy the \emph{Speaker Identification} sieve, or the \emph{Precise Constructs} sieve.
\item $\pi_j = attr$, if there is no mention $m_i, i < j$ satisfies the above two conditions.
\end{itemize}
Now, we can extend the generative model in Eq.~\ref{eq:model} to:
\begin{displaymath}
\begin{array}{rcl}
& & P(D, C) = P(D, C, \Pi) \\
 & = & \prod\limits_{j=1}^{n}P(m_j|m_{c_j}, \pi_j) P(c_j|\pi_j, j) P(\pi_j|j)
\end{array}
\end{displaymath}
where we define $P(\pi_j|j)$ to be uniform distribution. We model $P(m_j|m_{c_j}, \pi_j)$ and $P(c_j|\pi_j, j)$ in the following way:
\begin{displaymath}
\begin{array}{l}
P(m_j|m_{c_j}, \pi_j) = t(m_j|m_{c_j}, \pi_j) \\
P(c_j|\pi_j, j) = \left\{ \begin{array}{ll}
q(c_j|\pi_j, j) & \pi_j = attr \\
\frac{1}{j} & \textrm{otherwise}
\end{array}\right.
\end{array}
\end{displaymath}
where $\theta = \{t, q\}$ are parameters of our model. Note that in the attribute-matching mode ($\pi_j = attr$) we model $P(c_j|\pi_j, j)$ with parameter $q$, while in the other two modes, we use the uniform distribution. It makes sense because the position information is important for coreference resolved by matching attributes of two mentions such as resolving pronoun coreference, but not that important for those resolved by matching text or special relations like two mentions referring the same person and matching by the name.
\SetAlgoSkip{}
\begin{algorithm}[t]
{\small
\caption{Learning Model with EM} \label{al:learning}
\textbf{Initialization:} Initialize $\theta_0 = \{t_0, q_0\}$ \\
\For {$t=0$ {\bfseries to} $T$} {
    set all counts $c(\ldots) = 0$ \\
    \For {\textrm{each document} $D$} {
        \For {$j=1$ {\bfseries to} $n$} {
            \For {$k=0$ {\bfseries to} $j - 1$} {
                $L_{jk} = \frac{t(m_j|m_k,\pi_j)q(k|\pi_j, j)}{\sum\limits_{i = 0}^{j-1} t(m_j|m_i,\pi_j)q(i|\pi_j, j)}$ \\
                $c(m_j, m_k, \pi_j) \mathrel{+}= L_{jk}$ \\
                $c(m_k, \pi_j) \mathrel{+}= L_{jk}$ \\
                $c(k, j, \pi_j) \mathrel{+}= L_{jk}$ \\
                $c(j, \pi_j) \mathrel{+}= L_{jk}$
            }
        }
    }
    \tcp{\scriptsize{Recalculate the parameters}}
    $t(m|m', \pi) = \frac{c(m, m', \pi)}{c(m', \pi)}$ \\
    $q(k, j, \pi) = \frac{c(k, j, \pi)}{c(j, \pi)}$
}}
\end{algorithm}
\subsection{Features}
\label{subsec:feats}
In this section, we describe the features we use to represent mentions. Specifically, as shown in Table~\ref{tab:feats}, we use different features under different resolution modes. It should be noted that only the \emph{Distance} feature is designed for parameter $q$, all other features are designed for parameter $t$.
\begin{table}[t]
\centering
{\scriptsize
\begin{tabular}[t]{l|ccccc}
\hline
Corpora & \# Doc & \# Sent & \# Word & \# Entity & \# Mention \\
\hline
Gigaword & 3.6M & 75.4M & 1.6B & - & - \\
\hline
ON-Dev & 343 & 9,142 & 160K & 4,546 & 19,156 \\
\hline
ON-Test & 348 & 9,615 & 170K & 4,532 & 19,764 \\
\hline
\end{tabular}
}
\caption{Corpora statistics. ``ON-Dev'' and ``ON-Test'' are the development and testing sets of the OntoNotes corpus.}
\label{tab:data:stats}
\end{table}
\begin{table*}[t]
\centering
{\scriptsize
\bgroup
\def\arraystretch{1.5}
\begin{tabular}[t]{l|cccccc|cccccc}
 & \multicolumn{6}{c|}{\textbf{CoNLL'12 English development data}} 
 & \multicolumn{6}{c}{\textbf{CoNLL'12 English test data}} \\
\hline
 & MUC & B$^{3}$ & CEAF$_m$ & CEAF$_e$ & Blanc & CoNLL 
 & MUC & B$^{3}$ & CEAF$_m$ & CEAF$_e$ & Blanc & CoNLL \\
\hline
MIR & 65.39 & 54.89 & -- & 51.36 & -- & 57.21 & 64.64 & 52.52 & -- & 49.11 & -- & 55.42 \\
Stanford & 64.96 & 54.49 & 59.39 & 51.24 & 56.03 & 56.90 
& 64.71 & 52.26 & 56.01 & 49.32 & 53.92 & 55.43 \\
Multigraph & 66.22 & 56.41 & 60.87 & 52.61 & 58.15 & 58.41 
& 65.41 & 54.38 & 58.60 & 50.21 & 56.03 & 56.67 \\
\textbf{Our Model} & \textbf{67.89} & \textbf{57.83} & \textbf{62.11} & \textbf{53.76} & \textbf{60.58} & \textbf{59.83}
& \textbf{67.69} & \textbf{55.86} & \textbf{59.66} & \textbf{51.75} & 
\textbf{57.78} & \textbf{58.44} \\
\hdashline
IMS & 67.15 & 55.19 & 58.86 & 50.94 & 56.22 & 57.76 
& 67.58 & 54.47 & 58.17 & 50.21 & 55.41 & 57.42 \\
Latent-Tree & 69.46 & 57.83 & -- & 54.43 & -- & 60.57 
& 70.51 & 57.58 & -- & 53.86 & -- & 60.65 \\
Berkeley & 70.44 & 59.10 & -- & 55.57 & -- & 61.71
& 70.62 & 58.20 & -- & 54.80 & -- & 61.21 \\
LaSO & 70.74 & 60.03 & 65.01 & 56.80 & -- & 62.52 
& 70.72 & 58.58 & 63.45 & 59.40 & -- & 61.63 \\
Latent-Strc & 72.11 & 60.74 & -- & 57.72 & -- & 63.52 
& 72.17 & 59.58 & -- & 55.67 & -- & 62.47 \\
Model-Stack & 72.59 & 61.98 & -- & 57.58 & -- & 64.05 
& 72.59 & 60.44 & -- & 56.02 & -- & 63.02 \\
Non-Linear & 72.74 & 61.77 & -- & 58.63 & -- & 64.38 
& 72.60 & 60.52 & -- & 57.05 & -- & 63.39 \\
\hline
\end{tabular}
\egroup
}
\caption{F1 scores of different evaluation metrics for our model, together with two deterministic systems and one unsupervised system as baseline (above the dashed line) and seven supervised systems (below the dashed line) for comparison on CoNLL 2012 development and test datasets.}
\label{tab:comparison}
\end{table*}
\subsection{Model Learning}
For model learning, we run EM algorithm~\cite{dempster1977maximum} on our Model, treating $D$ as observed data and $C$ as latent variables. We run EM with 10 iterations and select the parameters achieving the best performance on the development data. Each iteration takes around 12 hours with 10 CPUs parallelly. The best parameters appear at around the 5th iteration, according to our experiments.The detailed derivation of the learning algorithm is shown in Appendix A. The pseudo-code is shown is Algorithm~\ref{al:learning}. We use uniform initialization for all the parameters in our model.

Several previous work has attempted to use EM for entity coreference resolution. \newcite{cherry-bergsma:2005} and \newcite{charniak-elsner:2009} applied EM for pronoun anaphora resolution. \newcite{ng:2008:EMNLP} probabilistically induced coreference partitions via EM clustering. Recently, \newcite{moosavi2014} proposed an unsupervised model utilizing the most informative relations and achieved competitive performance with the Stanford system.

\subsection{Mention Detection}\label{subsec:mention-detect}
The basic rules we used to detect mentions are similar to those of \newcite{Lee:2013:CL}, except that their system uses a set of filtering rules designed to discard instances of pleonastic \emph{it}, partitives, certain quantified noun phrases and other spurious mentions. Our system keeps partitives, quantified noun phrases and \emph{bare NP} mentions, but discards pleonastic \emph{it} and other spurious mentions.

\section{Experiments}
\label{sec:experiment}
\subsection{Experimental Setup}
\textbf{Datasets.} 
Due to the availability of readily parsed data, we select the APW and NYT sections of Gigaword Corpus (years 1994-2010) \cite{parker2011english} to train the model. Following previous work~\cite{chambers-jurafsky:2008:ACLMain}, we remove duplicated documents and the documents which include fewer than 3 sentences. The development and test data are the English data from the CoNLL-2012 shared task~\cite{pradhan-EtAl:2012:CoNLL-2012-ST}, which is derived from the OntoNotes corpus~\cite{hovy-EtAl:2006:HLT-NAACL06-Short}.
The corpora statistics are shown in Table~\ref{tab:data:stats}. Our system is evaluated with automatically extracted mentions on the version of the data with automatic preprocessing information (e.g., predicted parse trees).
\\
\textbf{Evaluation Metrics}.
We evaluate our model on three measures widely used in the literature: MUC~\cite{vilain1995model}, B$^{3}$~\cite{bagga1998algorithms}, and Entity-based CEAF (CEAF$_e$)~\cite{luo:2005:HLTEMNLP}. In addition, we also report results on another two popular metrics: Mention-based CEAF (CEAF$_m$) and BLANC~\cite{recasens2011blanc}. All the results are given by the latest version of CoNLL-2012 scorer~\footnote{\url{http://conll.cemantix.org/2012/software.html}}

\subsection{Results and Comparison}
Table~\ref{tab:comparison} illustrates the results of our model together as baseline with two deterministic systems, namely \textbf{Stanford}: the Stanford system~\cite{lee-EtAl:2011:CoNLL-ST} and \textbf{Multigraph}: the unsupervised multigraph system~\cite{martschat:2013:SRW}, and one unsupervised system, namely \textbf{MIR}: the unsupervised system using most informative relations~\cite{moosavi2014}. Our model outperforms the three baseline systems on all the evaluation metrics. Specifically, our model achieves improvements of 2.93\% and 3.01\% on CoNLL F1 score over the Stanford system, the winner of the CoNLL 2011 shared task, on the CoNLL 2012 development and test sets, respectively.
The improvements on CoNLL F1 score over the Multigraph model are 1.41\% and 1.77\% on the development and test sets, respectively. Comparing with the MIR model, we obtain significant improvements of 2.62\% and 3.02\% on CoNLL F1 score.

To make a thorough empirical comparison with previous studies, Table~\ref{tab:comparison} (below the dashed line) also shows the results of some state-of-the-art supervised coreference resolution systems --- \textbf{IMS}: the second best system in the CoNLL 2012 shared task~\cite{bjorkelund-farkas:2012:CoNLL-2012-ST}; \textbf{Latent-Tree}: the latent tree model~\cite{fernandes-dossantos-milidiu:2012:CoNLL-2012-ST} obtaining the best results in the shared task; \textbf{Berkeley}: the Berkeley system with the final feature set~\cite{durrett-klein:2013:EMNLP}; 
\textbf{LaSO}: the structured perceptron system with non-local features~\cite{bjorkelund-kuhn:2014:P14-1}; \textbf{Latent-Strc}: the latent structure system~\cite{TACL604}; \textbf{Model-Stack}: the entity-centric system with model stacking~\cite{clark-manning:2015:ACL-IJCNLP}; and \textbf{Non-Linear}: the non-linear mention-ranking model with feature representations~\cite{wiseman-EtAl:2015:ACL-IJCNLP}. Our unsupervised ranking model outperforms the supervised IMS system by 1.02\% on the CoNLL F1 score, and achieves competitive performance with the latent tree model. Moreover, our approach considerably narrows the gap to other supervised systems listed in Table~\ref{tab:comparison}.

\section{Conclusion}
We proposed a new generative, unsupervised ranking model for entity coreference resolution into which we introduced resolution mode variables to distinguish mentions resolved by different categories of information. Experimental results on the data from CoNLL-2012 shared task show that our system significantly improves the accuracy on different evaluation metrics over the baseline systems.

One possible direction for future work is to differentiate more resolution modes. Another one is to add more precise or even event-based features to improve the model's performance.

\section*{Acknowledgements}
This research was supported in part by DARPA grant FA8750-12-2-0342 funded under the DEFT program. Any opinions, findings,
and conclusions or recommendations expressed in this material are those of the authors and do not necessarily reflect the views of DARPA.

\bibliography{naaclhlt2016}
\bibliographystyle{naaclhlt2016}
{\bf Appendix A. Derivation of Model Learning} \\
 Formally, we iteratively estimate the model parameters $\theta$, employing the following EM algorithm:
\begin{description}
\item[E-step:] Compute the posterior probabilities of $C$, $P(C|D; \theta)$, based on the current $\theta$.
\item[M-step:] Calculate the new $\theta'$ that maximizes the expected complete log likelihood, $E_{P(C|D; \theta)}[\log P(D, C; \theta')]$
\end{description}
For simplicity, we denote:
\begin{displaymath}
{\small
\begin{array}{rcl}
P(C|D; \theta) & = & \tilde{P}(C|D) \\
P(C|D; \theta') & = & P(C|D)
\end{array}}
\end{displaymath}
In addition, we use $\tau(\pi_j|j)$ to denote the probability $P(\pi_j|j)$ which is uniform distribution in our model. Moreover, we use the following notation for convenience:
\begin{displaymath}
{\small
\theta(m_j, m_k, j, k, \pi_j) = t(m_j|m_k, \pi_j) q(k|\pi_j, j) \tau(\pi_j|j)
}
\end{displaymath}
Then, we have
\begin{displaymath}
{\scriptsize
{\setlength\arraycolsep{2pt}
\begin{array}{rl}
 & E_{\tilde{P}(c|D)} [\log P(D, C)] \\
= & \sum\limits_{C} \tilde{P}(C|D) \log P(D, C) \\
= & \sum\limits_{C} \tilde{P}(C|D) \big(\sum\limits_{j=1}^{n} \log t(m_j|m_{c_j}, \pi_j) + \log q(c_j|\pi_j, j) + \log \tau(\pi_j|j) \big) \\
= & \sum\limits_{j=1}^{n} \sum\limits_{k=0}^{j-1} L_{jk} \big(\log t(m_j|m_k, \pi_j) + \log q(k|\pi_j, j) + \log  \tau(\pi_j|j) \big)
\end{array}}}
\end{displaymath}
Then the parameters $t$ and $q$ that maximize $E_{\tilde{P}(c|D)} [\log P(D, C)]$ satisfy that
\begin{displaymath}
{\small
\begin{array}{rcl}
t(m_j|m_k, \pi_j) & = & \frac{L_{jk}}{\sum\limits_{i = 1}^{n} L_{ik}} \\
q(k|\pi_j, j) & = & \frac{L_{jk}}{\sum\limits_{i = 0}^{j-1} L_{ji}}
\end{array}}
\end{displaymath}
where $L_{jk}$ can be calculated by
\begin{displaymath}
{\small
\begin{array}{rcl}
L_{jk} & = & \sum\limits_{C, c_j=k} \tilde{P}(C|D) = \frac{\sum\limits_{C, c_j=k} \tilde{P}(C, D)}{\sum\limits_{C} \tilde{P}(C, D)} \\
 & = & \frac{\sum\limits_{C, c_j=k}\prod\limits_{i = 1}^{n}\tilde{\theta}(m_i, m_{c_i}, c_i, i, \pi_i)}{\sum\limits_{C}\prod\limits_{i = 1}^{n}\tilde{\theta}(m_i, m_{c_i}, c_i, i, \pi_i)} \\
 & = & \frac{\tilde{\theta}(m_j, m_k, k, j, \pi_j)\sum\limits_{C(-j)}\tilde{P}(C(-j)|D)}{\sum\limits_{i=0}^{j-1}\tilde{\theta}(m_j, m_i, i, j, \pi_j)\sum\limits_{C(-j)}\tilde{P}(C(-j)|D)} \\
 & = & \frac{\tilde{\theta}(m_j, m_k, k, j, \pi_j)}{\sum\limits_{i=0}^{j-1}\tilde{\theta}(m_j, m_i, i, j, \pi_j)} \\
 & = & \frac{\tilde{t}(m_j|m_k, \pi_j) \tilde{q}(k|\pi_j, j) \tilde{\tau}(\pi_j|j)}{\sum\limits_{i=0}^{j-1}\tilde{t}(m_j|m_i, \pi_j) \tilde{q}(i|\pi_j, j) \tilde{\tau}(\pi_j|j)} \\
 & = & \frac{\tilde{t}(m_j|m_k, \pi_j) \tilde{q}(k|\pi_j, j)}{\sum\limits_{i=0}^{j-1}\tilde{t}(m_j|m_i, \pi_j) \tilde{q}(i|\pi_j, j)}
\end{array}}
\end{displaymath}
where $C(-j) = \{c_1, \ldots, c_{j-1}, c_{j+1}, \ldots, c_{n}\}$. The above derivations correspond to the learning algorithm in Algorithm~\ref{al:learning}.
\end{document}